\pdfoutput=1

\documentclass[11pt]{article}

\usepackage[preprint]{acl}

\usepackage{times}
\usepackage{latexsym}
\usepackage{graphicx}

\usepackage[T1]{fontenc}
\usepackage{amsfonts}
\usepackage{amssymb}

\usepackage[utf8]{inputenc}

\usepackage{microtype}

\usepackage{inconsolata}

\usepackage{graphicx}

\usepackage{booktabs}
\usepackage{subfiles}
\usepackage{algorithm}
\usepackage{algorithmic}
\usepackage{enumitem}
\usepackage{amsmath}
\usepackage{tabularx}

\newcommand{\BenchName}{Mol-Hallu}

\title{How to Detect and Defeat Molecular Mirage: A Metric-Driven Benchmark for Hallucination in LLM-based Molecular Comprehension}

\author{{\small Hao Li$^{1,2,3,*}$, Liuzhenghao Lv$^{1,*}$, He Cao$^{2,4,*}$, Zijing Liu$^2$, Zhiyuan Yan$^{1}$, Yu Wang$^{1}$, Yonghong Tian$^{1,3,\dagger}$, Yu Li$^{2,\dagger}$, Li Yuan$^{1,3,\dagger}$} \\ $^1$Shenzhen Graduate School, Peking University \\ $^2$International Digital Economy Academy (IDEA) \\ $^3$Pengcheng Laboratory, Shenzhen, China \\ $^4$Hong Kong University of Science and Technology}

\begin{document}
\maketitle

\begin{abstract}


Large language models are increasingly used in scientific domains, especially for molecular understanding and analysis. However, existing models are affected by hallucination issues, resulting in errors in drug design and utilization. In this paper, we first analyze the sources of hallucination in LLMs for molecular comprehension tasks, specifically the knowledge shortcut phenomenon observed in the PubChem dataset. To evaluate hallucination in molecular comprehension tasks with computational efficiency, we introduce \textbf{Mol-Hallu}, a novel free-form evaluation metric that quantifies the degree of hallucination based on the scientific entailment relationship between generated text and actual molecular properties. Utilizing the Mol-Hallu metric, we reassess and analyze the extent of hallucination in various LLMs performing molecular comprehension tasks. Furthermore, the Hallucination Reduction Post-processing stage~(HRPP) is proposed to alleviate molecular hallucinations, Experiments show the effectiveness of HRPP on decoder-only and encoder-decoder molecular LLMs. Our findings provide critical insights into mitigating hallucination and improving the reliability of LLMs in scientific applications. \footnote{ lihao1984@pku.edu.cn, $^{\dagger}$ corresponding authors}


\end{abstract}
\section{Introduction}

\begin{figure}[!ht]
    \centering
    \includegraphics[width=1.0\linewidth]{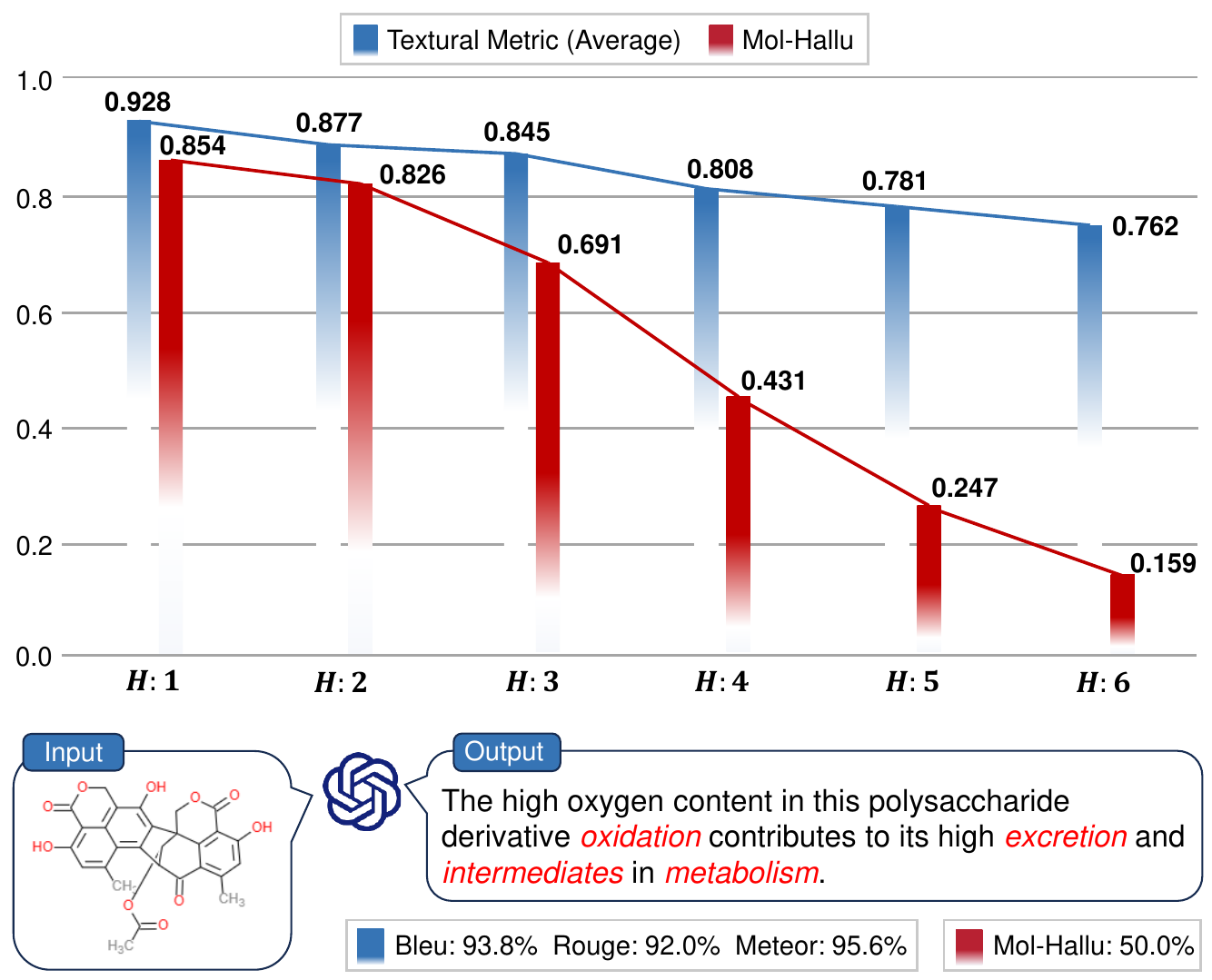}
    \vspace{-0.3in}
    \caption{
    (1)
    The top figure shows the scoring curves of \BenchName{} v.s. traditional metrics~(BLEU, ROUGE, METEOR) across varying degrees of hallucination. ${H}:n$ indicates that samples contain $n$ counterfactual errors, \text{\BenchName{}} imposes an exponential penalty on hallucination errors in text., whereas traditional metrics fail to evaluate biochemical hallucination in texts reasonably.
    (2) The bottom figure proposes a biochemical sample that suffers severe hallucination~(\textcolor{red}{red} are counterfactual entities) as an example. \BenchName{} precisely reflects the hallucination degree in scientific texts compared to traditional metrics. 
    }
    \label{fig:hallu-metric}
    \vspace{-1em}
\end{figure}

Large language models~(LLMs) are regarded as foundation models in scientific fields due to their outstanding cross-domain generalization capability~\cite{scientificllmsurvey,lv2024prollama}. In chemistry, LLMs are used for molecular property prediction~\cite{lv2024navigating,qian2023largelanguagemodelsempower} and molecular design~\cite{flam2022language,grisoni2023chemical}. These models bridge the gap between molecular structural and property features and the natural language descriptions, facilitating multiple chemical applications including virtual screening, drug design, retrosynthesis planning, etc.

Although LLMs have shown powering generation capability in biochemistry domains, they suffer from hallucinations~\cite{bang2023multitask} which leads to the fabrication of non-existent facts or inappropriate molecular properties~\cite{yao2023llm}. 
Hallucinations often arise when new biochemical knowledge introduced during the supervised fine-tuning~(SFT) stage conflicts with the model's pretrained knowledge~\cite{gekhman-etal-2024-fine}. The risky SFT strategy is frequently employed in various molecular LLMs~\cite{pei-etal-2023-biot5,fang2023mol,Llasmol}, demonstrating the ubiquity of hallucinations.

Several studies on molecular LLMs analyze the hallucination phenomenon in molecule comprehension tasks. MoleculeQA~\cite{lu2024moleculeqa} and MoleculeTextQA~\cite{laghuvarapumoltextqa} construct multi-choice QA datasets to assess the hallucination issues in molecular LLMs. However, these approaches require additional datasets for fine-tuning in the context of fixed-form evaluation~\cite{li2024can} and their multiple-choice question format is ill-suited for assessing the open-ended generation capabilities of large language models~\cite{wang2023fake}. 
To address this limitation, there is an urgent need for a free-form evaluation metric to quantify the degree of hallucination in molecular LLMs. Moreover, existing research has not yet analyzed the sources of hallucination in molecular LLMs or explored how to effectively mitigate these hallucinations.

To alleviate these issues, we first analyze the source of hallucinations in molecular LLMs and propose \textbf{\BenchName}, the first free-form evaluation metric specifically designed to assess hallucination. Our investigation focuses on the PubChemQA dataset~\cite{3dmolm}, a widely recognized benchmark source from PubChem database~\cite{wang2009pubchem} that aligns molecular structures with textual descriptions. We identify that knowledge shortcuts in this dataset hinder the alignment between molecular structures and biochemical entities, resulting in increased hallucinations. To quantify the extent of hallucinations, \BenchName{} leverages the union of the answer and the molecular general description, rewarding correct biomedical entities. 
The union and intersection are computed using an entailment model to determine whether the molecular descriptions entail a given text n-gram. 
To enhance evaluation, we curated a chemical entity database by automatically annotating PubChem and ChEMBL~\cite{chembl} datasets, to accurately retrieve biomedical entities from predicted texts. Fig.\ref{fig:hallu-metric} demonstrates the rationality of \BenchName{} for hallucination evaluation compared to traditional metrics including BLEU~\cite{bleu}, ROUGE~\cite{rouge}, and METEOR~\cite{meteor}.


To mitigate the hallucination in current molecular LLMs, we propose the Hallucination Reduction Post-processing~(HRPP) stage, which constructs a hallucination-sensitive preference dataset by leveraging our chemical entity database, thereby optimizing the accuracy of scientific entities in text generated by molecular LLMs. 
The HRPP approach has validated its effectiveness and generalizability under decoder-only and encoder-decoder language models, two basic paradigms of molecular LLMs.
Our contributions are summarized as follows:
\begin{itemize}[leftmargin=*]
\item We dive into the molecular hallucination issue and identify that bio-knowledge shortcuts in the dataset exacerbate LLM hallucination.
\item To measure the hallucination in molecular comprehension with efficiency, we propose the first free-form evaluation metric, \BenchName{}, which calculates the F1-score of scientific entities using entailment probability.
\item We further propose the hallucination reduction post-processing stage to alleviate the molecular hallucination using the hallucination-sensitive preference dataset.
\end{itemize}

\section{Related Works}

\subsection{LLMs for Molecular Comprehension}
Large language models pretrained with biochemical scientific data have shown substantial success in molecular comprehension tasks~\cite{feng2024bioactivity}. The molecular encoders capture 1D sequential features~\cite{chemformer, MolT5, fang2023mol, smilesbert}, 2D topological features~\cite{rong2020self, ying2021transformers, wang2022molecular}, and 3D structural patterns~\cite{GraphMVP, zhou2023unimol, Uni-Mol+} from the molecule. 
Related studies have adopted two primary strategies to bridge the heterogeneity gap between molecular and textual representations for enhanced comprehension. Firstly, the cross-modal contrastive learning strategy is applied to fine-tune molecular and textual encoders. MoMu~\cite{su2022molecular}, MoleculeSTM~\cite{liu2023multi}, and MolCA~\cite{liu2023molca} construct a joint representational space that aligns molecular features with their corresponding textual descriptions. As textual encoders grow in parameter size and inferential capability, some studies~\cite{cao-etal-2025-instructmol, cao-etal-2024-presto, omni-mol} have turned to supervised fine-tuning using molecular-text datasets to establish a pooling layer that maps molecular representations into the textual space of LLMs. However, constrained by the feature bias of molecular encoders and the prior knowledge of LLMs, current molecular LLMs are plagued by significant hallucination issues.

\subsection{Hallucination in Biochemical LLMs}

Alongside the advancement in reasoning, LLM models often generate nonsensical or unfaithful content to the provided source, referred as \textit{hallucination}~\cite{bang2023multitask, maynez-etal-2020-faithfulness}. The source-reference divergence phenomenon~\cite{ji2023survey} is the main cause of hallucination. The divergence comes from heuristic data collection~\cite{parikh2020totto} and imperfect representation learning during the training procedure~\cite{feng2020modeling} or erroneous decoding when conducting inference~\cite{dziri2021neural}. In molecular comprehension tasks, molecular LLMs often generate counterfactual content, which can lead to adverse consequences such as misleading users, and ultimately undermine the reliability of LLMs in scientific applications~\cite{lu2024moleculeqa}.




\begin{figure*}[!ht]
    \centering
    \includegraphics[width=1.0\linewidth]{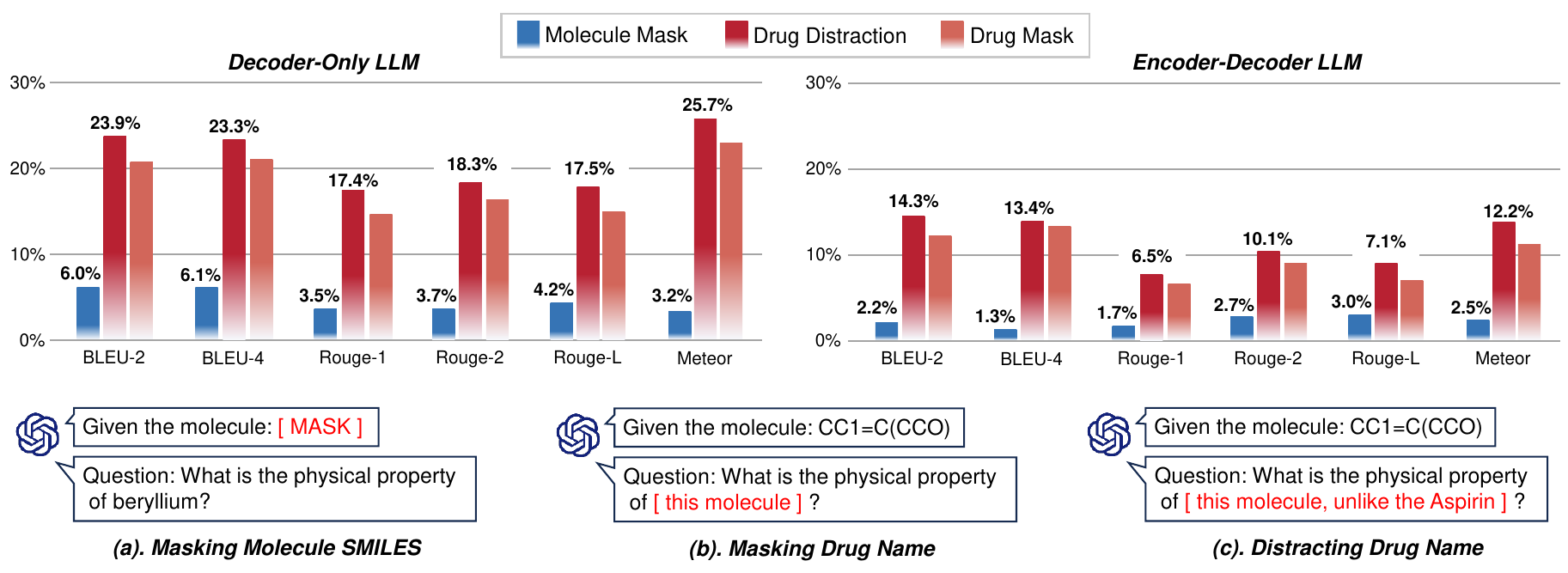}
    \vspace{-0.3in}
    \caption{Experiments demonstrate that in both decoder-only LLMs and encoder-decoder LLMs, molecule masking attacking has little impact while drug masking and distracting attackings lead to substantial decrease. 
    This indicates that the knowledge shortcut prompts LLMs to establish alignment between molecular properties and drug names instead of molecular structures, thereby deviating from the goal of molecular comprehension.
    }
    \label{fig:hallu_source}
    \vspace{-0.15in}
\end{figure*}

The evaluation of hallucinations in LLMs can be categorized into two main types: \textbf{(1) Fixed-form evaluation} and \textbf{(2) Free-form evaluation}. Fixed-form evaluation uses multi-choice QA datasets, such as MoleculeQA and MoleculeTextQA, to assess hallucinations. However, this method requires fine-tuning LLMs on hallucination datasets and uses a multi-choice format that differs from the open-ended nature of LLM tasks, making it less reflective of true hallucination extent. In contrast, free-form evaluation leverages automated functions for faster, more computationally efficient assessments. Hallucination detection methods also fall into two categories: \textbf{(1) Fact-checking-based methods}, which verify accuracy through external~\cite{factool, factscore} or internal knowledge~\cite{kadavath2022language, Chain-of-verification}, and \textbf{(2) Uncertainty estimation methods}~\cite{varshney2023stitch, manakul2023selfcheckgpt}, which detect hallucinations by quantifying model confidence without external references.
Our work bridges these approaches by introducing a free-form evaluation metric for molecular comprehension tasks. This method leverages ground truth while avoiding the need for external retrieval or fine-tuning, providing an efficient and domain-specific solution for hallucination detection. Currently, there are no such metrics for hallucination assessment in biochemical LLMs~\cite{rawte2023survey}, which limits the effectiveness of large scientific models in drug discovery. To address this, we propose the first free-form evaluation metric focused on the entailment of scientific entities, enabling more reliable application in this domain.

\section{Methodology}
In this section, we propose the definition, the source, the \BenchName{} evaluation metric, and the alleviation strategy for the molecular hallucination phenomenon.

\subsection{Definition of Molecular Hallucination}
\label{hallu_define}
Before delving into the source and evaluation of molecular hallucination, we first define the \textbf{Molecular Hallucination} as prediction texts that do not consist of the pharmacological or chemical properties of the molecule. Formally, given the molecule SMILES ${M}$ and the question $Q$. The hallucination is that LLM $f_\theta(\cdot)$ outputs non-existent or counterfactual scientific entities $E$ that do not satisfy the reality $\mathbb{T}$,
where $\mathbb{T}$ is the ground-truth entity set without any non-existent facts.


\subsection{Source of Molecular Hallucination}
The phenomenon of hallucination in LLMs arises from multiple sources, including inherent divergence and spurious noise within the data~\cite{lee-etal-2022-deduplicating}, as well as input knowledge bias~\cite{yin-etal-2023-large} in training paradigms during training and inference processes.

LLMs exhibit significant hallucinations in molecular comprehension tasks. Upon analyzing the PubChemQA dataset, we identified the \textbf{{bio-knowledge shortcuts}} exacerbate LLM hallucinations. 

\vspace{-5pt} 
\noindent\rule{\columnwidth}{0.4pt}
\vspace{-15pt} 
\begin{flushleft}
    \textsl{Molecule: Given a molecule} \textcolor{black}{\textbf{[SMILES]}}. \\
    \textsl{Question: What is the role of} \textcolor{black}{\textbf{[Drug Name]}} \textsl{in cellular processes?} 
\end{flushleft}
\vspace{-6pt} 
\noindent\rule{\columnwidth}{0.4pt}

To be more specific, bio-knowledge shortcuts refer to instances where drug names (e.g., beryllium) are present in molecular-related questions, leading the model to establish mappings between drug names and their physicochemical properties during supervised fine-tuning, rather than between molecular structures from SMILES and physicochemical properties, which is the original intent of molecular comprehension tasks. The existence of such shortcuts makes LLMs prone to hallucination due to changes or the absence of drug names and hinders their ability to infer physicochemical properties for novel molecules.

To prove this, we conduct attacks on the drug names contained in the questions within the molecular question-answer samples from the PubchemQA dataset and analyze the sources of hallucinations by observing the changes in hallucinations corresponding to different attack strategies~\cite{cao2024guide}. Specifically, given a sample and its corresponding question $Q$, we replace the drug name $D_j$ in $Q$ with (1) a masked pronoun $\mathrm{[\ this\ molecule\,]}$ and (2) a distracting drug name $\mathrm{[\ unlike\ D_{j}\,]}$. Fig.~\ref{fig:hallu_source} shows that two classes of commonly used scientific LLMs, the decoder-only models (e.g., Llama~\cite{llama2, dubey2024llama}) and the encoder-decoder models (e.g., T5~\cite{T5}), both exhibit severe hallucination phenomena~(-21\% Acc.) under two attack strategies. However, the absence of SMILES input has little influence on both models~(-5\% Acc.). This indicates that the models rely more on textual cues (e.g., drug names) than on SMILES structural information to infer molecular properties, highlighting their inability to align SMILES with molecular properties. This limits their generalization and reasoning capabilities for accurate molecular question-answering.


\subsection{\BenchName{} Metric}
To better quantify hallucination in LLMs for molecular comprehension tasks, we introduce the \textbf{\BenchName{}} evaluation metric to assess the extent of hallucination. This metric calculates Recall and Precision by comparing the entity entailment probability between the predicted answer $A_i$, the ground-truth answer $G_i$, and the molecular description $T_i$ corresponding to the molecule $M_i$, thereby evaluating the hallucination rate.

\subsubsection{Entity Entailment Probability} 
We define molecular hallucination as the phenomenon of scientific entity mismatches between predicted text and reference answers in Sec.~\ref{hallu_define}. To annotate scientific entities in the text, we employed Meta-llama-3.2~\cite{dubey2024llama} with a 10-shot prompting approach to automatically label scientific entities in captions and QA texts from the PubChemQA dataset and the ChEMBL dataset. After filtering based on inclusiveness, length, and semantics, we go through the human evaluation and obtain 97,219 chemical entities as the entity database. The statistic visualization below shows that half of the entities in our entity database are molecular structural entities, while the entities related to drug application, property, and natural source are balanced.
\begin{table}[h]
\footnotesize
    \centering
    \begin{tabular}{c|cccc}
        \toprule
        Type & Application & Property & Source & Structure\\
        \midrule
        Rate & 14.3\% & 19.7\% &12.0\% & 51.2\% \\
        \bottomrule
    \end{tabular}
    \label{tab:my_label}
\end{table}
Then, we introduce the entity entailment probability, defined as the probability that the presence of entity list $e$ is correct given the associated molecular descriptions and answers. Inspired by previous entailment works~\cite{dagan2005pascal}, we find that simple models are effective for entailment probability measurement. Here we apply the probability function as $w(\cdot)$, 
\begin{equation}
    w(e) = \sum_{j=1}^n \ \mathbf{1}(e_j \in \bar{\mathbb{T}}) / n,
\end{equation}
where $\mathbf{1}$ is the indicator function, $n$ is the entity number of $e$, and $\bar{\mathbb{T}}$ represents the set of all the entities present in description $T$. 
Then we compute the precision and the recall of the predicted text.

\begin{figure*}[!ht]
    \centering
    \includegraphics[width=1.0\linewidth]{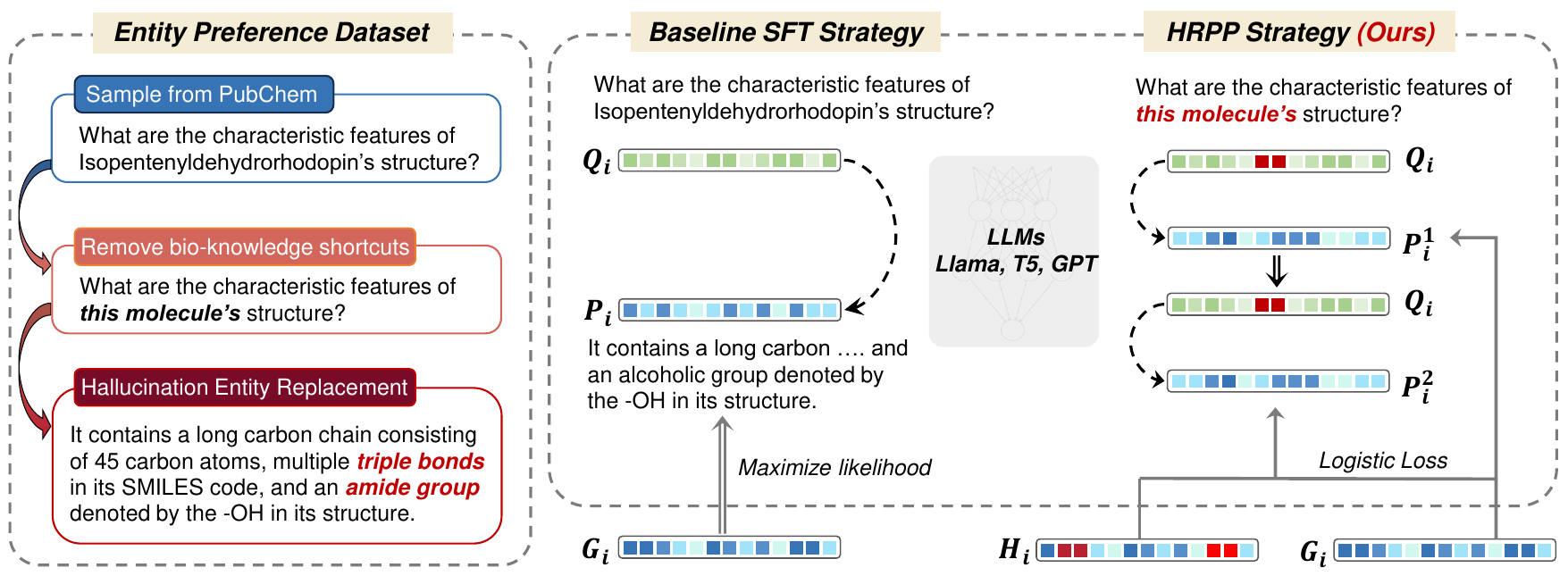}
    \vspace{-0.3in}
    \caption{The pipeline of entity preference dataset and our hallucination-reduction post-processing stage. The entity preference dataset is generated by removing bio-knowledge shortcuts and replacing entities with hallucinations. Then we apply the entity preference dataset for scientific-entity hallucination alleviation during the HRPP stage.}
    \label{fig:framework}
    \vspace{-0.1in}
\end{figure*}

\subsubsection{Entailed Precision} 
The entailed precision aims to represent the correct fraction of the n-gram entities in $mathbb{A}_i$, where $mathbb{A}_i$ is the set of all entities in predicted answer $A_i$. An n-gram entity $e$ is treated as correct if it appears in the ground-truth answer or if it appears in the molecular description, which is also a substantial correct answer. We apply $w(e)$ as the reward weight of the second scenario.

{\small
\begin{equation}
    P_e^{\text{n-gram}} = \sum_{e\in \mathbb{A}_i}[\mathrm{Pr}(e\in \mathbb{G}^\text{n-gram})+w(e)\mathrm{Pr}(e\notin \mathbb{G}^\text{n-gram})],
\end{equation}
}
Specifically, $P_e^{\text{n-gram}}$ represents the reward of the n-gram entity $e$. It receives a score of 1 if the ground-truth answer entails it. Otherwise, it receives a score of $w(e)$ if $e$ appears in the molecular description. We consider the numerator during the weight calculation of $P_e^{\text{n-gram}}$. Finally, we apply the geometric average to calculate the precision of the total sample group,
\begin{equation}
    \bar{P_e} = \mathrm{exp}(\sum_{\text{n-gram}=1}^{4}\frac{1}{4}\ \mathrm{log}\ P_e^\text{n-gram}),
\end{equation}
where we select the n-gram order from 1-4 as other metrics~\cite{papineni2002bleu,post2018call,dhingra2019handling}. Meanwhile, we calculate the n-gram matching score $\bar{P_\varnothing}$ for non-entity words. To balance the precision $\bar{P_e}$ from scientific entities and $\bar{P_\varnothing}$ from non-entities, we use the entity error count $\gamma$ as a weighting factor,
\begin{gather}
    \gamma = 1 - (N_\text{wrong} / N_\text{total})^{0.5}, \\
    \text{P} = \gamma \bar{P_\varnothing} + (1-\gamma) \bar{P_e},
\end{gather}
where $N_\text{wrong}$ and $N_\text{total}$ are wrong entity and total entity counts. $\text{P}$ represents the final precision score.

\begin{table*}[h]
\small
\centering
\renewcommand{\arraystretch}{1.2}  
\setlength{\tabcolsep}{2.5mm}        
\begin{tabular}{l|c|cccccc}
    \toprule
     \textbf{Models}& \textbf{\# Params} & \textbf{BLEU-2} & \textbf{BLEU-4} & \textbf{ROUGE-1} & \textbf{ROUGE-L} & \textbf{METEOR} & \textbf{\BenchName{}}$\uparrow$ \\
    \noalign{\hrule height 1.0pt}
    \multicolumn{8}{c}{{\textsl{Molecular-LLMs}}} \\
    \hline
    MolT5-small & 80M & 49.46 & 41.94 & 55.04 & 51.56 & 55.40 & 59.01 \\
    MolT5-base & 250M & 50.21 & 42.53 & \textbf{55.70} & \textbf{52.07} & {56.00} & 44.74 \\
    MolT5-large & 800M & 49.58 & 41.97 & 55.52 & 51.85 & 55.80 & 60.13 \\
    MoMu-small & 82M & 50.81 & 42.54 & 52.78 & 51.18 & 55.94 & 55.73 \\
    MoMu-base & 252M & 51.07 & 43.29 & 53.71 & 50.98 & 55.59 & \textbf{56.29} \\
    BioT5-base & 252M & 43.36 & 35.10 & 51.05 & 47.16 & 51.55 & 55.21 \\
    MolCA & 1.3B & \textbf{51.93} & \textbf{44.28} &55.00 & 51.41 & \textbf{56.79} & 55.82 \\
    3D-MoLM & 7B & 32.00 & 26.17 & 40.13 & 34.64 & 52.15 & 53.18 \\
    BioMedGPT & 10B & 37.31 & 31.29 & 39.62 & 36.87 & 48.31 & 43.88 \\
    \noalign{\hrule height 1.0pt}
    \multicolumn{8}{c}{{\textsl{General-LLMs}}} \\
    \hline
    T5-small & 60M & 49.97 & 42.40 & 54.88 & 51.16 & 55.47 & 59.07 \\
    T5-base & 220M & 51.01 & 43.27 & 55.89 & 52.17 & \textbf{56.43} & 60.21 \\
    T5-large & 770M & 50.79 & 42.85 & \textbf{55.98} & \textbf{52.23} & 56.42 & \textbf{60.93} \\
    Llama-2 & 7B & 28.15 &  23.24 &  35.14 & 30.41 &  46.87 & 53.78\\
    Llama-3.1 & 8B & \textbf{52.19} & \textbf{43.51} & {55.41} & {51.18}& {57.48} & 60.14\\
    \noalign{\hrule height 1.0pt}
    \multicolumn{8}{c}{{\textsl{Universal-LLM-API (Few-shot)}}} \\
    \hline
    Qwen-2.5-Instruct & 32B & 35.72&27.51&43.59& 38.22&49.63 & 49.97 \\
    Qwen-Reason~(QwQ) & 32B &18.62&13.62&27.33&23.32&35.14& 25.61 \\
    DeepSeek-V3 & 671B & \textbf{49.31} & 39.86 & \textbf{53.96} & \textbf{48.37} & \textbf{57.69} & \textbf{62.16} \\
    DeepSeek-R1 & 671B & 32.12 & 24.17 & 41.77 & 37.56 & 40.65 & 46.65 \\
    GPT-4o-\footnotesize{20241120} & 1.8T & 47.78 & \textbf{41.74} & 51.97 & 46.99 & 51.24 & 55.71 \\
    o1-mini & 300B & 40.22 &31.06&46.99 &41.81 &51.88 & 51.23 \\
    \bottomrule
\end{tabular}
\caption{
 Experimental results for hallucination evaluation across molecular LLMs (fine-tuned), general LLMs (fine-tuned), and universal LLMs (API-based inference). We report accuracy (\%) using both standard textual metrics and our proposed hallucination-specific evaluation metric.
}
\label{main_results}
\end{table*}

\subsubsection{Entailed Recall}
The entailed recall $\text{R}$ reflects the extent to which the model misses correct words. $\text{R}$ is computed between predicted $A$ and ground truth $G$ to ensure that entities and other n-gram words with high frequency in the ground truth receive a higher score when predicted correctly. We also apply the geometric average to get $\text{R}$ from $\text{R}_{1...n}$. 

\subsubsection{Smoothing \& Combination}
\BenchName{} employs the geometric average to compute entailed precision due to its ability to reflect compound changes accurately. However, when a component approaches 0, the geometric average also tends to 0. To mitigate this issue, we apply smoothing $\theta\text{=}10^{-5}$ to components close to 0.
After the precision smoothing, we calculate the F1-score based on the entailed precision $\text{P}$ and recall $\text{R}$. 

{\small
\begin{gather}
    \mathrm{Mol\text{-}Hallu}(A, G, T) = 2\text{P}\cdot \text{R} / (\text{P}+\text{R}), \\
    \mathrm{Mol\text{-}Hallu}(f_\theta)=\frac{1}{N}\sum_{i=1}^{N}\mathrm{Mol\text{-}Hallu}(A_i, G_i, T_i),
\end{gather}}
where the F1-scores from all samples generated by the model $f_\theta$ are arithmetic averaged to represent the hallucination rate of $f_\theta$.

\subsection{Hallucination Reduction Post-processing}
To mitigate the hallucination in LLM-based molecular comprehension, we propose the Hallucination Reduction Post-processing~(HRPP) stage. As shown in Fig.~\ref{fig:framework}, HRPP consists of two main steps: (1) reducing the model's reliance on entity name shortcuts through supervised fine-tuning, and (2) improving response accuracy and reducing hallucination using Direct Preference Optimization~(DPO) with a hallucination-sensitive preference dataset.

To mitigate the model’s tendency to generate hallucinated responses due to over-reliance on entity name shortcuts, we employ a supervised fine-tuning approach. Given a training dataset \(\mathcal{D} = \{(q_i, G_i)\}_{i=1}^{N}\), where \(Q_i\) is the input text and \(G_i\) is the corresponding ground truth response, we preprocess \(Q_i\) by masking entity names, replacing them with "this molecule" to prevent shortcut learning. We then optimize the model parameters \(\theta\) by minimizing the cross-entropy loss:
\begin{equation}
    \mathcal{L}_{\text{CE}}(\theta) = - \sum_{i=1}^{N} \sum_{t=1}^{T} \log P_{\theta}(G_i^t\ |\ Q_i, G_i^{<t})
\end{equation}
where \(T\) is the sequence length, \(N\) is the sample number, and \(P_{\theta}\) represents the model’s probability distribution over the vocabulary.

To further improve response accuracy and factual consistency of molecular LLMs, we first construct a hallucination-sensitive preference dataset \(\mathcal{D}_p = \{(q_i, G_i^{\text{+}}, G_i^{\text{-}})\}_{i=1}^{M}\), where \(G_i^{\text{+}}\) represents the preferred response, and \(G_i^{\text{-}}\) represents the less preferred response. As shown in Fig.~\ref{fig:framework} left, to construct this dataset, we randomly extract 2000 QA pairs from the training set. The ground truth \(G_i\) is designated as \(G_i^{\text{+}}\). To generate the negative sample~\(G_i^{\text{-}}\), we introduce entity perturbations by randomly replacing certain entities in \(G_i\) with different ones using our chemical entity database. Additionally, we sample four responses from the model at a high temperature for each $q_i$, incorporating them into the set of \(G_i^{\text{-}}\) responses.

We use DPO to optimize the model by maximizing the divergence between the likelihood of preferred and rejected responses:
\begin{equation}
    \mathcal{L}_{\text{}}(\theta)= 
    -\sum_{i=1}^{M} \log \sigma \left( 
    \beta  \log 
    \frac{ P_{\theta}(G_i^{\text{+}} | q_i) P_{\text{r}}(G_i^{\text{-}} | q_i) }
    { P_{\theta}(G_i^{\text{-}} | q_i) P_{\text{r}}(G_i^{\text{+}} | q_i) }
    \right) 
\end{equation}
where \(\sigma(\cdot)\) is the sigmoid function, \( P_{\text{r}} \) is the reference model, and \(\beta\) is a temperature hyperparameter that controls the strength of preference learning. In the experiment section, we apply HRPP to decoder-only LLMs and encoder-decoder LLMs for effectiveness analysis.

\section{Experiments}

\begin{table*}[ht]
\small
\centering
\renewcommand{\arraystretch}{1.35}  
\setlength{\tabcolsep}{2.5mm}        
\begin{tabular}{l|cccccc|c}
    \toprule
     \textbf{Molecular LLMs}& {BLEU-2} & {BLEU-4} & {ROUGE-1} & {ROUGE-2} & {ROUGE-L} & {METEOR} & \textbf{\BenchName{}}$\uparrow$\\
    \noalign{\hrule height 0.3pt}
    \hline
    \textbf{MolT5} & 34.48 & 26.54 & 45.13 & 28.17 & 41.34 & 37.08 & 46.15 \\
    \quad + HRPP & 40.65 & 30.73 & 47.47 & 29.98 & 43.54 & 44.31 & \textbf{49.03} \\
    \hline
    \textbf{Llama-3.1-8B} & 33.18 & 24.75 & 44.19 & 27.12 & 40.66 & 37.57 & 44.21 \\
    \quad + HRPP & 38.79 & 28.95 & 46.12 & 28.41 & 42.17 & 43.27 & \textbf{46.28} \\
    \bottomrule
\end{tabular}
\vspace{-0.05in}
\caption{
 Hallucination Reduction Post-processing~(HRPP) has substantial improvements in textural metrics and our \BenchName{} metric, demonstrating its effectiveness on both decoder-only models and encoder-decoder-based models. 
}
\label{tab:hallu_improve}
\vspace{-0.1in}
\end{table*}

\begin{figure*}[h]
    \centering
    \includegraphics[width=0.8\linewidth]{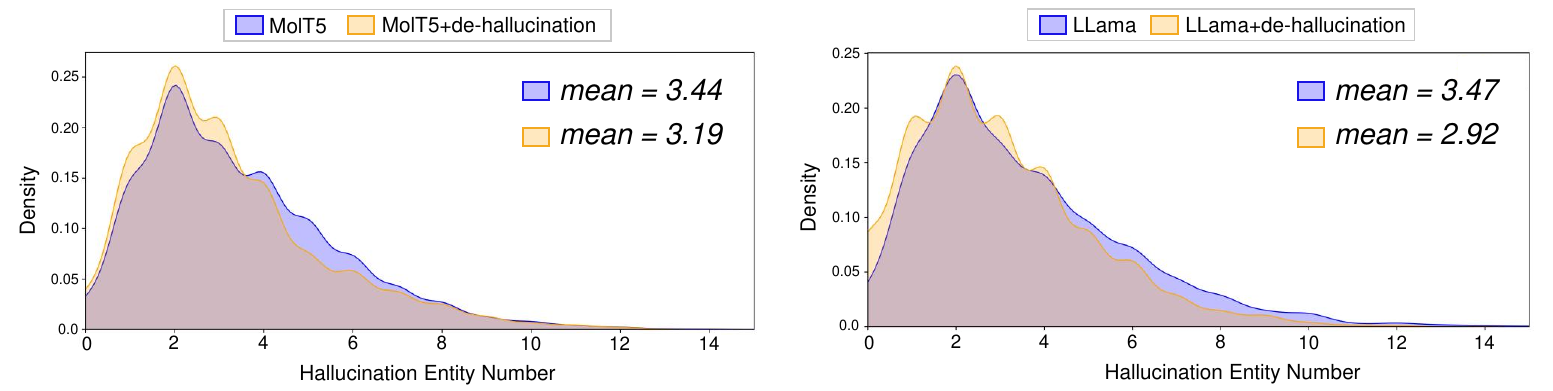}
    \vspace{-10pt}
    \caption{\textbf{Hallucination Distribution Comparison}. We visualize the distributions of hallucination entity numbers between molecular LLMs~(MolT5, Llama-3.1) and their de-hallucination versions. Our HRPP effectively mitigates the frequent occurrence of hallucinations in cases, shifting the distribution peak closer to 0.}
    \label{fig:ablation_hallu_num}
    \vspace{-0.1in}
\end{figure*}

\subsection{Baseline Models and Training Procedures}

To comprehensively evaluate the LLM performance in molecular conprehension, we introduce three categories of LLMs as baselines, including scientifically fine-tuned LLMs, general-purpose LLMs, and commercial LLMs. Specifically, LLMs fine-tuned with biochemical knowledge exhibit strong capabilities in modeling molecular SMILES and protein sequences. We evaluate their hallucination levels on the PubChemQA dataset in a zero-shot manner. General-purpose LLMs, trained extensively in natural scenarios, although less adept at modeling molecular SMILES compared to scientifically fine-tuned LLMs, possess stronger reasoning abilities. Commercial LLMs have stronger prior knowledge and reasoning capabilities due to their large parameter sizes. We conduct paid evaluations using the APIs of commercial LLMs, employing 10-shot instruction fine-tuning to generate responses to molecular-related queries.

\subsection{Main Results}
We summarize and analyze the baseline performances in Table.\ref{main_results}.

\textbf{Hallucinations in baseline models. }\ 
(1) The hallucination metric remains within the range of 40-60\%, with an average of 3-4 counterfactual entities present, indicating significant room for improvement.
(2) The degree of hallucination is not necessarily positively correlated with model performance. While MolT5-base shows comparable performance to MolT5-small and MolT5-large, its hallucination is notably more severe. In contrast, 3D-MoLM exhibits moderate performance but demonstrates a lower degree of hallucination.

\textbf{Structure Comparison: Encoder-Decoder v.s. Decoder-only.} Encoder-decoder models surpass other structures in molecular comprehension tasks due to their compact size and excellent performance. We observe that T5-based models, represented by T5-finetune, MolT5, and MoMu, exhibit strong performance on the MolecularQA task even in their small versions, surpassing molecular LLMs based on Llama by 2.7\% and GPT-4 by 13\%. This is attributed to the T5 model's encoder-decoder structure, which employs a span corruption pre-training strategy. Additionally, its smaller parameter count supports full-parameter fine-tuning instead of the LoRA fine-tuning used in Llama, resulting in better generalization in few-shot scenarios within the biochemistry domain.

\begin{table*}[ht]
\footnotesize
\centering
\begin{tabular}{cllll}
\toprule
    \textbf{Molecule} & \textbf{Query-Type} & \textbf{Ground truth} & \textbf{Our answer} & \textbf{Metric}\\
    \midrule
    \begin{tabular}[b]{c}
    \vspace{-1.8em}  
    \includegraphics[width=0.10\textwidth]{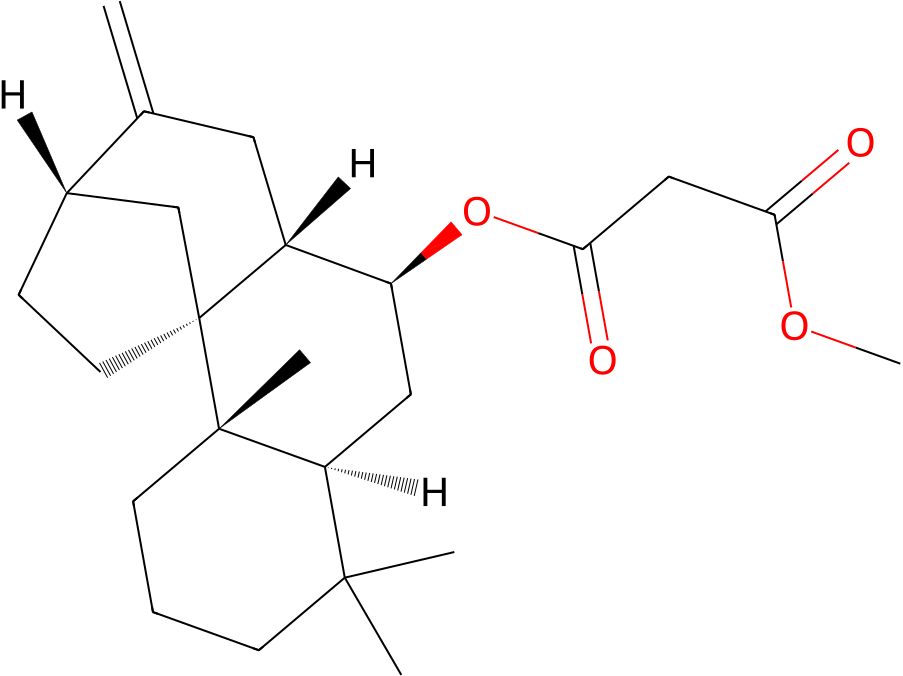} 
    \end{tabular} & 
    \begin{tabularx}{0.07\textwidth}{@{}X@{}}
    Isolated Area
    \end{tabularx} & 
    \begin{tabularx}{0.27\textwidth}{@{}X@{}}
    This compound is isolated from the plants Sorbus cuspidata and Calceolaria dentata.
    \end{tabularx} & 
    \begin{tabularx}{0.27\textwidth}{@{}X@{}}
    \textcolor{blue}{Hexaen} is isolated from the plants \textcolor{blue}{pentahydroxy and benzoate}.
    \end{tabularx} &
    \begin{tabularx}{0.10\textwidth}{@{}X@{}}
    B:~78.9\% \\
    R:~86.4\% \\
    M:~87.9\% \\
    \textcolor{red}{M-H}:~{43.3\%}
    \end{tabularx}
    \\

    \midrule
    \begin{tabular}[b]{c}
    \vspace{-1.8em}  
    \includegraphics[width=0.12\textwidth]{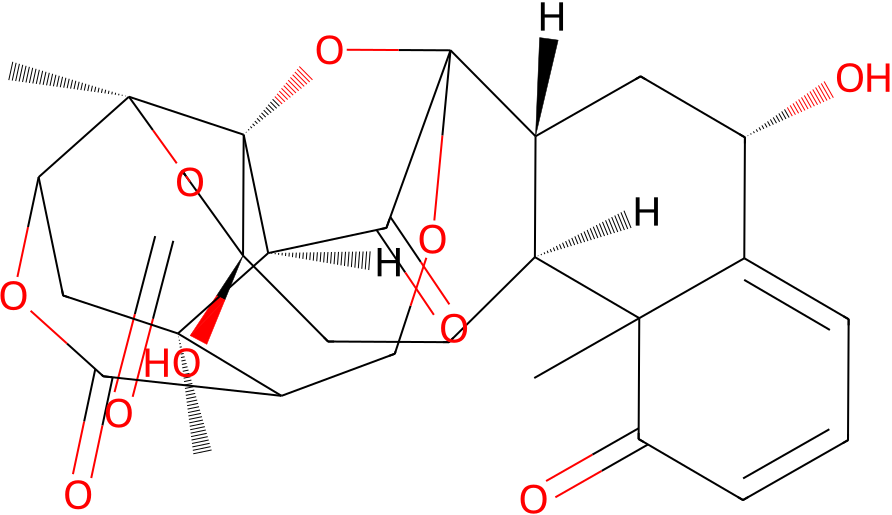} 
    \end{tabular} & 
    \begin{tabularx}{0.07\textwidth}{@{}X@{}}
    Potential Reactivity
    \end{tabularx} & 
    \begin{tabularx}{0.27\textwidth}{@{}X@{}}
    This compound has potential reactivity towards nucleophiles and bases due to the presence of ketone and lactone groups.
    \end{tabularx} & 
    \begin{tabularx}{0.27\textwidth}{@{}X@{}}
    This compound has potential reactivity towards \textcolor{blue}{aromaticity and methoxy} due to the presence of \textcolor{blue}{solubility and reactivity groups}.
    \end{tabularx} &
    \begin{tabularx}{0.10\textwidth}{@{}X@{}}
    B:~92.2\% \\
    R:~93.3\% \\
    M:~93.9\% \\
    \textcolor{red}{M-H}:~{66.1\%}
    \end{tabularx}
    \\
    
\bottomrule
\end{tabular}
\vspace{-0.05in}
\caption{\textbf{Case Studies for \BenchName{} and Other Textural Metrics}. Our \BenchName{} exhibits stronger sensitivity to hallucinated outputs under different question types in molecule comprehension.}
\vspace{-0.2in}
\label{tab:case_study}
\end{table*}

\textbf{Reward strategies in LLMs amplify hallucination.}\ 
Deepseek-R1 and o1-mini have widely adopted reinforcement learning as an effective approach to enhance the LLM reasoning capabilities for complex problems. However, this optimization strategy often leads to a hallucination increase~\cite{hallureasoning}. We observe a similar phenomenon in Molecular Comprehension. In the LLM-API part of Table.\ref{main_results}, we compare Qwen, Deepseek, and GPT-4, with their reasoning-enhanced versions on scientific QA tasks. The results indicate a significant decline in both prediction quality and factual accuracy, attributed to: (1) the trade-off between improved reasoning in math/code tasks and the reduced reliance on prior knowledge, making it harder to address scientific questions; and (2) the tendency of reasoning-enhanced LLMs to generate chain-of-thought outputs, which often contain more hallucinated entities. Therefore, balancing reasoning and hallucination in domain-specific scenarios remains a critical challenge.

\textbf{Extra protein knowledge: no benefit to hallucination.} \ 
During pretraining, extending the dataset to include both chemical molecules and protein macromolecules cannot alleviate hallucination. Instead, it leads to a decrease in performance for molecular understanding tasks. In Table~\ref{main_results}, BioMedGPT~\cite{luo2023biomedgpt} and BioT5 utilize various protein dataset size~(1.8M, 27M) as additional knowledge. However, their performance and hallucination assessment are inferior to the MolT5-based model due to the structural differences between FASTA-based protein inputs and SMILES-based molecular inputs, as well as the significant domain-specific entity differences between proteins and chemical molecules. Consequently, the incorporation of such knowledge fails to enhance generalization or reduce hallucination.

\subsection{Analysis for Hallucination Reduction}
In Table.~\ref{tab:hallu_improve} and Fig.~\ref{fig:ablation_hallu_num}, we dive into the hallucination reduction post-processing~(HRPP) and analyze its effectiveness on hallucination alleviation.

\noindent \textbf{Effectiveness of HRPP Stage.} \ Our HRPP stage shows effectiveness and generalizability on both decoder-only and T5-based models. Table.~\ref{tab:hallu_improve} shows that HRPP has substantial improvements for molecular LLMs, bringing an average of 4.0\% improvements on textural metrics. For the hallucination evaluation, our HRPP stage also achieves effective hallucination alleviation on both decoder-only structure~($2.9\%\uparrow$) and T5-based structure~($2.0\%\uparrow$). 
Meanwhile, we observed a significant improvement in the BLEU and METEOR~(5-7\%) during the HRPP stage, while the ROUGE series improvement is less pronounced~(1-2\%). This indicates that molecular LLMs optimized through HRPP tend to generate text with higher precision in scientific entities and more accurate semantics. However, missing scientific entities still occur in some answers due to the ROUGE series metrics being more sensitive to recall.

\noindent \textbf{Hallucination Distribution Analysis. }
To analyze the impact of HRPP on hallucinated samples generated by LLMs, we visualize the change in the number of counterfactual entities $N_c$ before and after the HRPP stage. In Fig.~\ref{fig:ablation_hallu_num}, HRPP effectively suppresses highly hallucinated samples~($N_c > 4$) in both decoder-only and encoder-decoder LLMs. After the HRPP stage, the distribution of counterfactual entities significantly shifts toward the low-hallucination region~($0<N_c<3$), demonstrating the efficacy of the HRPP stage.

\subsection{Case Studies}

We select samples with hallucinations and demonstrate a numerical comparison between our \BenchName{} metric and traditional textual metrics.
Table.~\ref{tab:case_study} shows that \BenchName{} are more sensitive to hallucinations. When the prediction and ground truth share similar sentence structures but differ in scientific entities, Mol-Hallu assigns a lower score, whereas traditional evaluation methods consider them semantically similar. Additional case studies are proposed in the Appendix.A1.


\section{Conclusion and Future Work}

In conclusion, our work aims to evaluate and alleviate the LLM's hallucination in molecular comprehension. By attacking the scientific entities in molecule-related questions, we identify the bio-knowledge shortcuts in the PubChem dataset as the hallucination source of the molecular comprehension task. We further propose the hallucination evaluation metric, \BenchName{}, for molecular comprehension. To alleviate the hallucination, we propose the hallucination reduction post-processing strategy with a molecular hallucination-sensitive preference dataset constructed based on entity replacement. Experimental results demonstrate that various LLM architectures significantly suppressed hallucinations with this strategy.

\clearpage
\section*{Limitations}
We conclude our limitations into the following aspects: (1) Our Mol-Hallu metric relies on a scientific entity database to localize scientific entities in predicted texts and evaluate the degree of hallucination. Although the current entity database demonstrates excellent coverage in the small molecule domain, its coverage in other scientific fields, such as protein understanding, remains limited. Future work should incorporate domain-specific terminologies to construct a more comprehensive entity database. (2) The current benchmark lacks full fine-tuning of large models due to insufficient training resources. Future efforts will focus on fine-tuning LLMs with 7B+ parameters and exploring the relationship between the performance and hallucination levels of molecular LLMs under scaling laws.

\section*{Potential Risks}
Although \BenchName{} provides a viable metric for hallucination assessment in the molecular comprehension domain, there remains a risk of abuse. \BenchName{} evaluation may not accurately represent a model's hallucination level over all chemistry-related scenarios.

\bibliography{related_papers}

\appendix

\section{Appendix}

\begin{table*}[ht]
\footnotesize
\centering
\begin{tabular}{cllll}
\toprule
    \textbf{Molecule} & \textbf{Q-Type} & \textbf{Ground truth} & \textbf{Our answer} & \textbf{Metric}\\

    \midrule
    \begin{tabular}[b]{c}
    \vspace{-1.8em}  
    \includegraphics[width=0.15\textwidth]{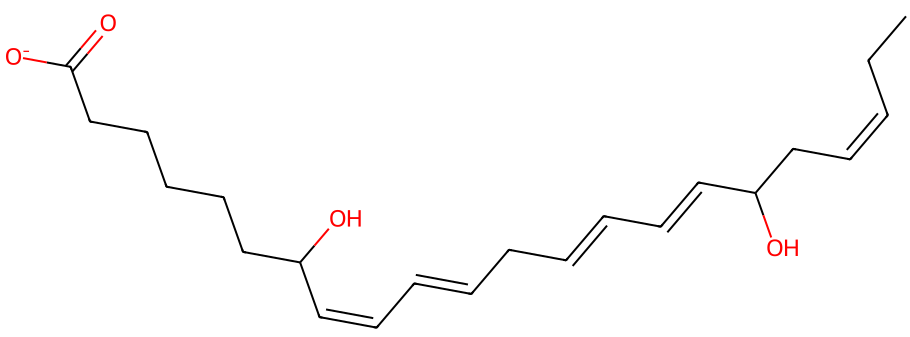} 
    \end{tabular} & 
    \begin{tabularx}{0.07\textwidth}{@{}X@{}}
    Structure
    \end{tabularx} & 
    \begin{tabularx}{0.27\textwidth}{@{}X@{}}
    It has a docosanoid structure with a 22-carbon chain and unsaturated bonds at positions 8, 15, and 19. It also contains hydroxyl groups at positions 7 and 17.
    \end{tabularx} & 
    \begin{tabularx}{0.27\textwidth}{@{}X@{}}
    It has a docosanoid structure with a 22-carbon chain and \textcolor{blue}{sedative at disorders} 8, 15, and 19. It also contains \textcolor{blue}{appetite at disorders} 7.
    \end{tabularx} &
    \begin{tabularx}{0.10\textwidth}{@{}X@{}}
    B:~88.4\% \\
    R:~87.5\% \\
    M:~91.5\% \\
    \textcolor{red}{M-H}:~{66.5\%}
    \end{tabularx}
    \\

    \midrule
    \begin{tabular}[b]{c}
    \vspace{-1.8em}  
    \includegraphics[width=0.15\textwidth]{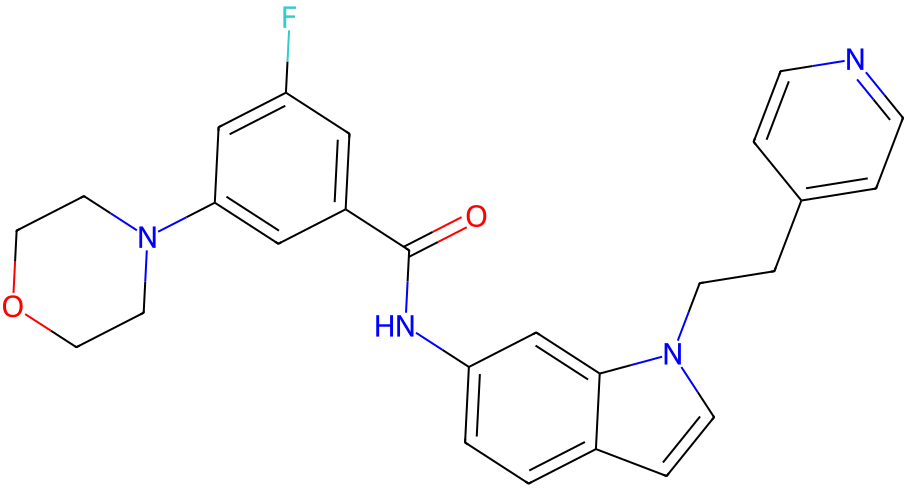} 
    \end{tabular} & 
    \begin{tabularx}{0.07\textwidth}{@{}X@{}}
    Class
    \end{tabularx} & 
    \begin{tabularx}{0.27\textwidth}{@{}X@{}}
    This organic compound belongs to the class of benzamides.
    \end{tabularx} & 
    \begin{tabularx}{0.27\textwidth}{@{}X@{}}
    This organic compound belongs to the class of \textcolor{blue}{carboxylic acid}.
    \end{tabularx} &
    \begin{tabularx}{0.10\textwidth}{@{}X@{}}
    B:~83.8\% \\
    R:~82.6\% \\
    M:~85.7\% \\
    \textcolor{red}{M-H}:~{1.6\%}
    \end{tabularx}
    \\

    \midrule
    \begin{tabular}[b]{c}
    \vspace{-1.8em}  
    \includegraphics[width=0.10\textwidth]{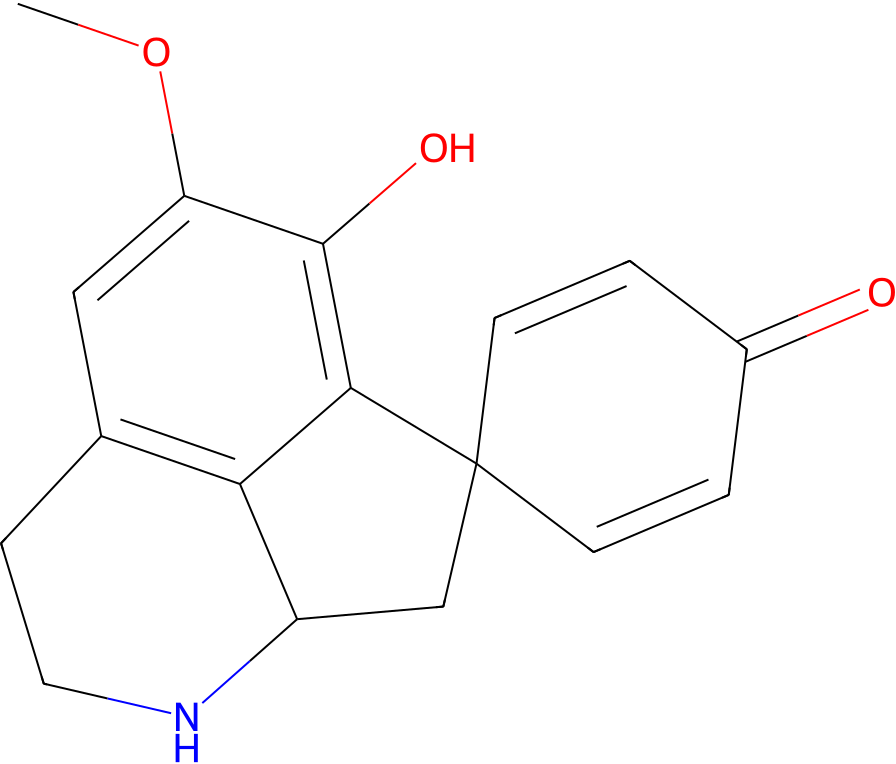} 
    \end{tabular} & 
    \begin{tabularx}{0.07\textwidth}{@{}X@{}}
    Solubility Property
    \end{tabularx} & 
    \begin{tabularx}{0.27\textwidth}{@{}X@{}}
    This molecule has solubility in both polar and nonpolar solvents due to the presence of a hydroxy group~(-OH) and a methoxy group~(-OCH3).
    \end{tabularx} & 
    \begin{tabularx}{0.27\textwidth}{@{}X@{}}
    This molecule has \textcolor{blue}{shaking} in both polar and \textcolor{blue}{insomnia} due to the presence of \textcolor{blue}{a hallucinations~(-OH) and a seizures~(-OCH3)}.
    \end{tabularx} &
    \begin{tabularx}{0.10\textwidth}{@{}X@{}}
    B:~88.3\% \\
    R:~87.9\% \\
    M:~90.9\% \\
    \textcolor{red}{M-H}:~{63.9\%}
    \end{tabularx}
    \\

    \midrule
    \begin{tabular}[b]{c}
    \vspace{-1.8em}  
    \includegraphics[width=0.10\textwidth]{Molecule/4.png} 
    \end{tabular} & 
    \begin{tabularx}{0.07\textwidth}{@{}X@{}}
    Isolated Area
    \end{tabularx} & 
    \begin{tabularx}{0.27\textwidth}{@{}X@{}}
    This compound is isolated from the plants Sorbus cuspidata and Calceolaria dentata.
    \end{tabularx} & 
    \begin{tabularx}{0.27\textwidth}{@{}X@{}}
    \textcolor{blue}{Hexaen} is isolated from the plants \textcolor{blue}{pentahydroxy and benzoate}.
    \end{tabularx} &
    \begin{tabularx}{0.10\textwidth}{@{}X@{}}
    B:~78.9\% \\
    R:~86.4\% \\
    M:~87.9\% \\
    \textcolor{red}{M-H}:~{43.3\%}
    \end{tabularx}
    \\

    \midrule
    \begin{tabular}[b]{c}
    \vspace{-1.8em}  
    \includegraphics[width=0.12\textwidth]{Molecule/5.png} 
    \end{tabular} & 
    \begin{tabularx}{0.07\textwidth}{@{}X@{}}
    Potential Reactivity
    \end{tabularx} & 
    \begin{tabularx}{0.27\textwidth}{@{}X@{}}
    This compound has potential reactivity towards nucleophiles and bases due to the presence of ketone and lactone groups.
    \end{tabularx} & 
    \begin{tabularx}{0.27\textwidth}{@{}X@{}}
    This compound has potential reactivity towards \textcolor{blue}{aromaticity and methoxy} due to the presence of \textcolor{blue}{solubility and reactivity groups}.
    \end{tabularx} &
    \begin{tabularx}{0.10\textwidth}{@{}X@{}}
    B:~92.2\% \\
    R:~93.3\% \\
    M:~93.9\% \\
    \textcolor{red}{M-H}:~{66.1\%}
    \end{tabularx}
    \\

    \midrule
    \begin{tabular}[b]{c}
    \vspace{-1.8em}  
    \includegraphics[width=0.17\textwidth]{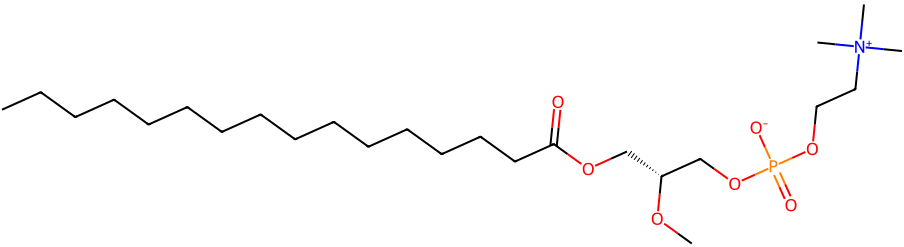} 
    \end{tabular} & 
    \begin{tabularx}{0.07\textwidth}{@{}X@{}}
    Structure
    \end{tabularx} & 
    \begin{tabularx}{0.27\textwidth}{@{}X@{}}
    The molecule has a glycerol backbone with a hexadecanoyl group attached to the sn-1 position and a methyl group attached to the sn-2 position. It also has a phosphate group and a choline molecule attached to the sn-3 position.
    \end{tabularx} & 
    \begin{tabularx}{0.27\textwidth}{@{}X@{}}
    The molecule has a glycerol backbone with a hexadecanoyl group attached to the sn-1 position and a methyl group attached to the \textcolor{blue}{PbSO4 position}. It also has a \textcolor{blue}{zinc group} and a \textcolor{blue}{silver} molecule attached to the \textcolor{blue}{copper position}.
    \end{tabularx} &
    \begin{tabularx}{0.10\textwidth}{@{}X@{}}
    B:~79.6\% \\
    R:~87.8\% \\
    M:~84.1\% \\
    \textcolor{red}{M-H}:~{67.9\%}
    \end{tabularx}
    \\

    \midrule
    \begin{tabular}[b]{c}
    \vspace{-1.8em}  
    \includegraphics[width=0.12\textwidth]{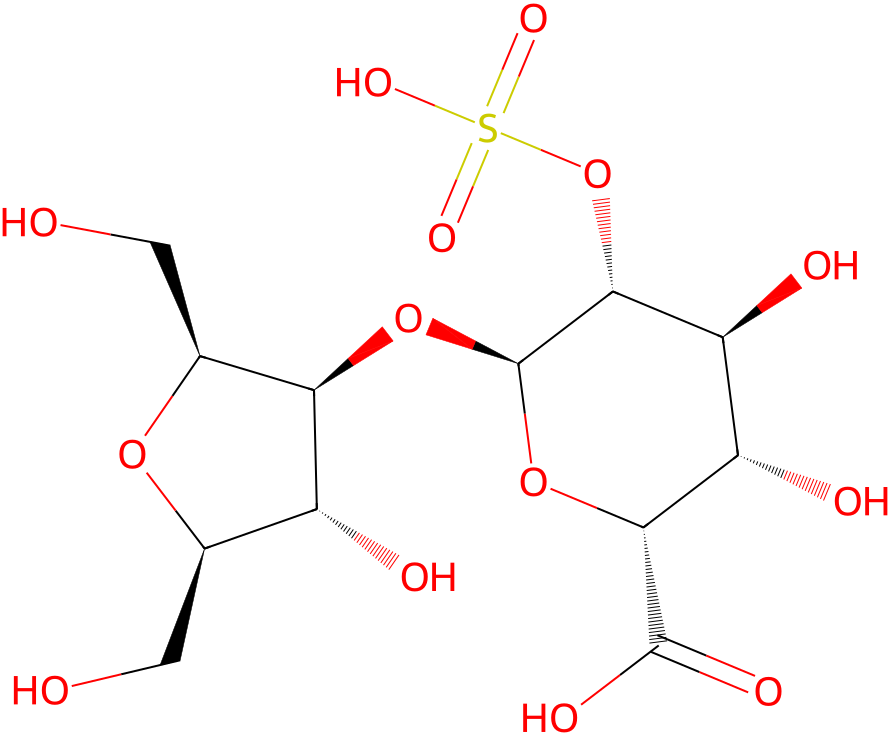} 
    \end{tabular} & 
    \begin{tabularx}{0.07\textwidth}{@{}X@{}}
    Chemical Classify
    \end{tabularx} & 
    \begin{tabularx}{0.27\textwidth}{@{}X@{}}
    The compound is classified as a carbohydrate acid derivative, meaning it is a derivative of a carboxylic acid that contains a carbohydrate moiety. It is also categorized as an oligosaccharide sulfate, indicating it is a sulfated oligosaccharide with multiple sugar units and sulfate groups.
    \end{tabularx} & 
    \begin{tabularx}{0.27\textwidth}{@{}X@{}}
    The compound is classified as a \textcolor{blue}{carbohydrate acid postganglionic}, meaning it is a postganglionic of a \textcolor{blue}{effector-cell acid} that contains a \textcolor{blue}{carbohydrate moiety}. It is also categorized as \textcolor{blue}{a receptor}, indicating it is a \textcolor{blue}{sulfated oligosaccharid}e with \textcolor{blue}{multiple muscle and sulfate bronchoconstriction}.
    \end{tabularx} &
    \begin{tabularx}{0.10\textwidth}{@{}X@{}}
    B:~78.1\% \\
    R:~86.2\% \\
    M:~85.2\% \\
    \textcolor{red}{M-H}:~{65.5\%}
    \end{tabularx}
    \\

\bottomrule
\end{tabular}
\caption{
Additional case studies for Mol-Hallu and other textural metrics. Our Mol-Hallu exhibits stronger sensitivity to
hallucinated outputs under different question types in molecule comprehension.
}
\label{tab:app_case_study}
\end{table*}

\subsection{Case Studies for PubchemQA Dataset}
We systematically enumerated samples with varying degrees of hallucination from the PubchemQA dataset and compared the scores of traditional metrics (BLEU-2/4, ROUGE-1/2/L, and METEOR) with those of \BenchName{}. 
Fig.~\ref{tab:app_case_study} provides 7 samples from PubchemQA, where Q-Type represents the question type of the sample, $B, R, M, M\text{-}H$ in Metric represents the average of BLEU-2/4, the average of Rouge-1/2/L, Meteor, and our \BenchName{} metric. The experiment results in Fig.~\ref{tab:app_case_study} covered diverse molecular structures and question types, demonstrating that \BenchName{} accurately reflects the hallucination degree across different scenarios, exhibiting robust performance and domain adaptability. Notably, in the second case, where the model's prediction completely deviated from the ground truth, \BenchName{} assigned a low score of 1.6\%, while traditional metrics, misled by superficial sentence similarities, provided significantly higher scores~(83.8\%, 87.5\%, 91.5\%). This contrast not only highlights the inherent limitations of traditional metrics in evaluating hallucinations in biochemical texts but also further validates the reliability and superiority of \BenchName{} in detecting semantic errors in scientific entities.

\subsection{The Evaluation Introduction}
In this subsection, we provide the detailed information for traditional textural evaluation metrics for LLM prediction in Question-Answering tasks.

\textbf{BLEU: } (Bilingual Evaluation Understudy) is a precision-based metric widely used for evaluating the quality of machine-generated text by comparing it to one or more reference texts. It measures the overlap of n-grams (typically up to 4-grams) between the generated text and the references. The BLEU score is calculated as follows:

\begin{equation}
\text{BLEU} = BP \cdot \exp\left(\sum_{n=1}^N w_n \log p_n\right)
\end{equation}

where \( BP \) is the brevity penalty to penalize short translations, \( w_n \) is the weight for each n-gram precision \( p_n \), and \( N \) is the maximum n-gram order (usually 4).

\textbf{ROUGE: } (Recall-Oriented Understudy for Gisting Evaluation) is a recall-oriented metric commonly used for evaluating summarization tasks. It measures the overlap of n-grams, word sequences, or word pairs between the generated text and the reference texts. The most frequently used variant, ROUGE-N, is defined as:

{\small
\begin{equation}
\text{ROUGE-N} = \frac{\sum_{\mathcal{R}} \sum_{\text{n-gram} \in \mathcal{R}} \text{C}_{\text{match}}(\text{n-gram})}{\sum_{\mathcal{R}} \sum_{\text{n-gram} \in \mathcal{R}} \text{C}(\text{n-gram})}
\end{equation}
}

where \( \text{C}_{\text{match}}(\text{n-gram}) \) is the number of n-grams co-occurring in both the generated and reference texts $\mathcal{R}$, and \( \text{C}(\text{n-gram}) \) is the total number of n-grams in the reference.

\textbf{METEOR: } (Metric for Evaluation of Translation with Explicit ORdering) is a metric designed to address some limitations of BLEU by incorporating synonymy, stemming, and word order. It calculates a weighted harmonic mean of precision and recall, with a penalty for word order discrepancies. The METEOR score is computed as:

\begin{equation}
\text{METEOR} = (1 - \gamma \cdot \text{Penalty}) \cdot \frac{10 \cdot P \cdot R}{R + 9 \cdot P}
\end{equation}

where \( P \) and \( R \) are precision and recall, respectively, \( \gamma \) is a parameter controlling the penalty weight, and \( \text{Penalty} \) is a function of the number of word order violations.

\subsection{Licenses and Terms of Use for Models and Datasets}
In this study, we employed multiple models and datasets, each subject to distinct licensing terms. The following is a summary of these licenses along with their respective usage conditions.

\textbf{MolT5}: Released by blender-nlp under the BSD 3-Clause License. This license permits free use, modification, and distribution, provided that specific conditions are met, such as retaining the copyright notice and disclaimer. Commercial use is allowed, but endorsement or promotion of derived products using the copyright holder’s name requires prior written permission. The license also includes a liability disclaimer, stating that the software is provided "as is" without warranties or guarantees.

\textbf{MoMu}: Released under the MIT License. This license permits free use, modification, and distribution, including for commercial purposes, as long as the original copyright notice and permission notice are retained. The software is provided "as is," without any warranties or guarantees, and the authors bear no liability for any claims, damages, or other issues arising from its use.

\textbf{BioT5}: Released under the MIT License. This license permits free use, modification, and distribution, including for commercial purposes, as long as the original copyright notice and permission notice are retained. The software is provided "as is," without any warranties or guarantees, and the authors bear no liability for any claims, damages, or other issues arising from its use.

\textbf{3D-MoLM}: Released under the Apache 2.0 License. This license permits free use, modification, and distribution, including for commercial purposes, provided that the original copyright notice and license terms are retained. Users are allowed to patent their modifications but must grant a license for any patented contributions. The software is provided "as is," without warranties or liabilities, and users must include a notice stating any modifications made to the original version.

\textbf{BioMedGPT}: Released under the MIT License. This license permits free use, modification, and distribution, including for commercial purposes, as long as the original copyright notice and permission notice are retained. The software is provided "as is," without any warranties or guarantees, and the authors bear no liability for any claims, damages, or other issues arising from its use.

\textbf{T5}: Released under the Apache 2.0 License. This license permits free use, modification, and distribution, including for commercial purposes, provided that the original copyright notice and license terms are retained. Users are allowed to patent their modifications but must grant a license for any patented contributions. The software is provided "as is," without warranties or liabilities, and users must include a notice stating any modifications made to the original version.

\textbf{Llama-2}: Released by Meta under the Llama 2 Community License. This license permits free use, modification, and distribution, but restricts the model’s use for training other language models and imposes specific conditions for commercial use, such as active user limits.

\textbf{Llama-3.1}: Released by Meta under the Llama 3.1 Community License. This license permits free use, modification, and distribution, with requirements such as attribution, compliance with Meta’s Acceptable Use Policy, and display of "Built with Llama" for derivative works. Commercial use is allowed, but entities with over 700 million monthly active users must obtain a separate license from Meta. The license includes disclaimers of warranty and liability, and any legal disputes fall under the jurisdiction of California law.

\textbf{Qwen-2.5-Instruct}~\cite{qwen2.5}: Released under the Apache 2.0 License. This license permits free use, modification, and distribution, including for commercial purposes, provided that the original copyright notice and license terms are retained. Users are allowed to patent their modifications but must grant a license for any patented contributions. The software is provided "as is," without warranties or liabilities, and users must include a notice stating any modifications made to the original version.

\textbf{Qwen-Reason~(QwQ)}~\cite{qwq-32b-preview}: Released under the Apache 2.0 License. This license permits free use, modification, and distribution, including for commercial purposes, provided that the original copyright notice and license terms are retained. Users are allowed to patent their modifications but must grant a license for any patented contributions. The software is provided "as is," without warranties or liabilities, and users must include a notice stating any modifications made to the original version.

\textbf{DeepSeek-V3}~\cite{liu2024deepseekv3}: Released by DeepSeek under the DeepSeek License (v1.0, Oct 23, 2023). It grants a free, global, irrevocable license for using, modifying, and distributing DeepSeek-V3, with strict restrictions on military use, harm, misinformation, discrimination, and unauthorized data processing. Users must enforce these limits in derivatives. DeepSeek may restrict misuse remotely and disclaims warranties and liability. Governed by Chinese law (PRC), jurisdiction in Hangzhou.

\textbf{DeepSeek-R1}~\cite{guo2025deepseekr1}: Released under the MIT License. This license permits free use, modification, and distribution, including for commercial purposes, as long as the original copyright notice and permission notice are retained. The software is provided "as is," without any warranties or guarantees, and the authors bear no liability for any claims, damages, or other issues arising from its use.

\textbf{GPT-4o-20241120}: Released by OpenAI. It is proprietary software. Access to this model is provided through OpenAI's platforms, such as ChatGPT and the Azure OpenAI Service, under specific subscription plans. The model is not open-source and is subject to OpenAI's terms of service and usage policies.

\textbf{o1-mini}:  Released by OpenAI. It is proprietary software. Access to o1-mini is provided through OpenAI's API and platforms, such as ChatGPT, under specific subscription plans. The model is not open-source and is subject to OpenAI's terms of service and usage policies.

\textbf{PubChemQA~(3D-MoIT)}: Released under the Apache 2.0 License. This license permits free use, modification, and distribution, including for commercial purposes, provided that the original copyright notice and license terms are retained. Users are allowed to patent their modifications but must grant a license for any patented contributions. The software is provided "as is," without warranties or liabilities, and users must include a notice stating any modifications made to the original version.

\textbf{ChEMBL}: Released under the Creative Commons Attribution-ShareAlike 3.0 Unported License. This license allows free use, modification, and distribution of the dataset, but requires appropriate attribution and mandates that any derivative works or modifications must be distributed under the same license.

\section{Hallucination Analysis beyond Scientific Domain}
Beyond textual and chemical domain hallucinations~\cite{li2025decoupled}, the phenomenon has been systematically studied in multimodal domains, including 3D vision and video-text understanding. In 3D object generation tasks~\cite{tang2024cycle3d}, hallucination frequently manifests as geometrically implausible structures or physically inconsistent depth relationships. Recent work by \cite{yu2024evagaussians,feng2025ae, zhang2025repaint123} proposed a depth-aware generation framework that incorporates geometric priors from LiDAR point clouds during the diffusion process, effectively suppressing structural hallucinations in ShapeNet benchmarks through explicit depth constraint optimization.

The multimodal domain~\cite{li2024freestyleret,jia2025uni} presents unique hallucination challenges due to temporal dynamics and cross-modal alignment requirements. \cite{li2023tg,jin2023diffusionret,li2022joint} identified that hallucinations in standard video captioning systems stem from misaligned fine-grained representations between visual frames and textual descriptors. Their solution introduced a hierarchical contrastive learning mechanism with three-level alignment: (1) global scene matching, (2) object-state verification, and (3) temporal relation distillation. This approach reduced cross-modal hallucination by 37\% on ActivityNet. Also, the proposed method in \cite{li2023weakly}achieves hallucination reduction by injecting depth prior knowledge, utilizing point cloud data as conditional inputs to regulate the model's spatial perception.

These findings suggest that explicit structural priors (in 3D domains) and reinforced cross-modal alignment (in video-text understanding) constitute effective hallucination suppression strategies that may inspire analogous solutions for chemical LLMs.

\end{document}